\documentclass{vgtc}                          % final (conference style)
\onlineid{TBD}

\vgtccategory{Empirical Study}

\title{Benchmarking Multimodal Large Language Models for \\ Scientific Visualization Literacy}

\author{Patrick Phuoc Do\thanks{e-mail: mdo23@nd.edu} %
\and Chau M. Ta\thanks{e-mail: cta@nd.edu} %
\and Chaoli Wang\thanks{e-mail: chaoli.wang@nd.edu}}
\affiliation{\scriptsize University of Notre Dame}

\abstract{%
Multimodal large language models (MLLMs) are increasingly used to interpret visualizations, yet current evaluations remain largely chart-centric and provide limited evidence of understanding of scientific visualization (SciVis). We benchmark six MLLMs on the scientific visualization literacy assessment test, a standardized SciVis literacy assessment comprising 49 items based on 18 scientific visualizations and illustrations, spanning 8 techniques and 11 task types. We evaluate three closed-source and three open-source models under a closed-world protocol and compare their performance using data from 485 human participants. Results show that current MLLMs do not exhibit uniform SciVis literacy. Gemini is the strongest model overall, exceeding the human mean across the evaluated subsets, whereas the open-source models remain below the human baseline. Performance is highly uneven across techniques and tasks: models perform best on scientific illustration, search, and spatial understanding, but struggle on texture-based and integration-based visualizations and on quantitative estimation. Error analysis reveals recurring failures in fine-grained quantitative estimation, flow-direction interpretation, and grounded encoding interpretation. These findings position SciVis literacy as a necessary benchmark dimension for evaluating multimodal AI systems. 
\hot{Our code and model outputs are publicly available at \url{https://github.com/patdmp/mllm-scivis-lit-benchmark}}.
}


\graphicspath{{figures/}{pictures/}{images/}{./}} % where to search for the images

\usepackage{times}                     % we use Times as the main font
\usepackage{mathptmx}                  % use matching math font
\usepackage{graphicx}
\usepackage{algorithm}
\usepackage{algorithmic}
\usepackage{amsmath}
\usepackage{comment}
\usepackage{caption}
\usepackage{url}
\usepackage[pagebackref,bookmarks]{hyperref}
\usepackage{booktabs}
\usepackage{multirow}
\usepackage{makecell}
\usepackage{tabularx}
\usepackage{float}
\usepackage{accessibility}

\usepackage{listings}
\newcommand{\hot}[1]{{\color{black} #1}}
\begin{document}

%% The ``\maketitle'' command must be the first command after the
%% ``\begin{document}'' command. It prepares and prints the title block.
%% the only exception to this rule is the \firstsection command
\vspace{-0.1in}
\firstsection{Introduction}
\maketitle
Scientific visualization (SciVis) has long served as an important medium for exploring, analyzing, and communicating complex scientific data, including simulation output, medical imaging, climate data, and other spatially grounded scientific phenomena \cite{mccormick1987visualization, Kehrer-TVCG13}. In contrast to information visualization (InfoVis), SciVis typically represents data with intrinsic spatial or physical structure. This distinction is closely tied to the notions of physically based data and given spatialization, which differentiate much of SciVis from InfoVis \cite{Tory-InfoVis04}. In addition, when interpreting SciVis, viewers need to reason about spatial structure, multivariate relationships, and time-varying or physically meaningful patterns encoded in the visual representation \cite{Kehrer-TVCG13, Tory-InfoVis04}.

Meanwhile, multimodal large language models (MLLMs) have rapidly advanced in their ability to process images alongside text and to support visualization interpretation, scientific communication, and visual analytics workflows \cite{achiam2023gpt, team2023gemini, Zhao-TVCG25}. However, existing evidence suggests that their performance on visualization tasks remains uneven, and current evaluations have focused primarily on InfoVis rather than SciVis \cite{bendeck2024empirical, Hong-TVCG25}. As a result, whether current MLLMs possess \emph{SciVis literacy} remains an open question \cite{Wang-EVIS26}.

To address this gap, we benchmark MLLMs using the scientific visualization literacy assessment test (SVLAT) \cite{Do-arXiv26}, a recently introduced instrument designed specifically to measure how well non-experts read, understand, and interpret scientific visualizations and illustrations. Because SVLAT spans multiple SciVis techniques and tasks under a closed-world design, it provides a suitable benchmark for evaluating SciVis understanding in MLLMs.

Using SVLAT, our goal is to examine whether current MLLMs demonstrate measurable SciVis literacy and, if so, where their strengths and weaknesses lie across techniques and task categories. More broadly, we position SciVis literacy as a missing benchmark dimension for multimodal AI systems. By doing so, we aim to contribute a clearer evaluation target for future work on trustworthy model assessment, model-assisted scientific communication, and human--AI interaction around scientific visual content.

\vspace{-0.1in}
\section{Related Work}
{\bf Visualization literacy assessment.}\
Visualization literacy refers to a viewer's ability to read, understand, and interpret data visualizations~\cite{borner2019data, maltese2015data, beschi2025characterizing}. Boy et al.~\cite{Boy-TVCG14} introduced a principled approach based on item response theory (IRT) to assess visualization literacy, showing that literacy can be modeled as a latent ability rather than treated solely as raw accuracy. Building on this psychometric perspective, Lee et al.~\cite{Lee-TVCG17} developed VLAT, a standardized assessment for non-expert users that measures literacy across 12 visualization types and 8 visualization tasks. CALVI~\cite{ge2023calvi} extended the notion of visualization literacy beyond routine reading by focusing on critical thinking about misleading or erroneous visualizations. To improve practicality and reduce administrative time, later work introduced shorter, more specialized instruments. For instance, Mini-VLAT~\cite{pandey2023mini} condensed VLAT into a brief but effective short-form test while preserving much of its measurement utility. Adaptive versions of VLAT and CALVI~\cite{Cui-TVCG24} reduced testing burden while retaining reliability and validity. These studies show that visualization literacy is a mature topic for assessment within InfoVis.

{\bf SciVis literacy.}\
Compared with InfoVis, SciVis introduces additional interpretive demands. Scientific visualizations often encode data with inherent spatial and physical structure, requiring users to reason about geometry, depth, topology, motion, and relationships among scalar, vector, or tensor fields. Prior work on SciVis tasks has emphasized the importance of task taxonomies tailored to scientific data analysis, especially for volume data and spatial reasoning~\cite{laha2015classification}. These characteristics make SciVis literacy qualitatively different from InfoVis literacy in the context of conventional charts and graphs. Despite the importance of SciVis in scientific analysis and communication, standardized assessment in this area has remained largely absent until recently. SVLAT~\cite{Do-arXiv26} is a recently proposed psychometrically grounded instrument for measuring SciVis literacy. SVLAT is designed around 8 canonical SciVis techniques and 11 interpretation tasks, enabling literacy to be measured in a way that better reflects the demands of scientific visual representations. Because SVLAT is standardized, validated, and explicitly constructed for scientific visual material, it provides an appropriate foundation for benchmarking MLLMs on SciVis literacy rather than relying on chart-oriented proxies.

{\bf MLLMs and visualization understanding.}\
Work on machine understanding of visualized data initially developed through benchmark datasets for chart and figure question answering. Early datasets such as FigureQA~\cite{Kahou-ICLRW18} and DVQA~\cite{Kafle-CVPR18} focused on synthetic scientific-style figures and bar charts, respectively, establishing controlled testbeds for visual reasoning over plotted data. PlotQA~\cite{Methani-WACV20} moved closer to realistic scientific plots by introducing open-vocabulary and numerical answers, while ChartQA~\cite{Masry-ACL22} expanded the task to charts requiring both visual and logical reasoning. More recent benchmarks have further stressed the limitations of current MLLMs. CharXiv~\cite{Wang-NeurIPS24} evaluated realistic charts collected from scientific papers and showed that prior benchmarks can overestimate model capability, and MultiChartQA~\cite{Zhu-ACL25} studied multi-chart reasoning, where models must integrate information across several related visualizations. Beyond chart-only settings, SPIQA~\cite{Pramanick-NeurIPS24} evaluated multimodal question answering over figures and tables embedded in scientific papers, extending the problem toward document-grounded scientific figure understanding.

Within the visualization community, recent work has begun to study MLLMs through the lens of visualization literacy itself. Bendeck and Stasko~\cite{bendeck2024empirical} evaluated GPT-4 on a suite of visualization-literacy-related tasks and found a mixed pattern of strengths and weaknesses: the model could identify higher-level trends and extreme values, but it struggled with precise value retrieval, color discrimination, and consistency. Hong et al.~\cite{Hong-TVCG25} then examined whether LLMs truly possess visualization literacy by testing them on modified VLAT-style visualizations designed to probe generalization and reliance on visual evidence rather than prior knowledge. Their findings suggest that model performance can degrade substantially under benign visual modifications, raising concerns about robustness and genuine visual grounding. Das et al.~\cite{Das-TVCG25} further showed that structured prompting can substantially improve performance on VLAT-style tasks, highlighting that measured capability depends not only on the model itself but also on the evaluation protocol. Complementing these studies, Varona et al.~\cite{varona2026mllms} analyzed visualization-literacy barriers in MLLMs by qualitatively coding erroneous responses, providing a taxonomy of failure modes that helps explain why models struggle even when benchmark accuracy appears competitive. Dong and Crisan~\cite{Dong-TVCG25} further probed the internal reasoning behavior of vision-language models for chart understanding, arguing that answer correctness alone is insufficient for characterizing visualization literacy in models.

%However, this literature remains largely chart-centric. 
Most existing datasets, evaluations, and prompting studies focus on InfoVis-style charts or chart-like figures rather than SciVis-native representations. Together, these studies show that evaluation of MLLMs on visualization tasks remains largely chart-centric, leaving SciVis-native assessment comparatively underexplored.

\vspace{-0.1in}
\section{Methodology}
{\bf Assessment instrument.}\
We evaluate MLLMs using SVLAT \cite{Do-arXiv26}, a standardized assessment of SciVis literacy. SVLAT consists of 49 items based on 18 scientific visualizations and illustrations, including static images and animations, spanning 8 SciVis techniques and 11 task types. It follows a closed-world design in which each item is intended to be answerable using only the provided visualization and caption, making it suitable for assessing SciVis literacy in MLLMs while limiting reliance on external knowledge.

{\bf Model selection and configuration.}\
We evaluate both closed-source and open-source MLLMs. The closed-source models are GPT-5.4 (GPT)\cite{gpt5.4}, Claude-Opus-4.6 (Claude)\cite{claudeOpus4.6}, and Gemini-3.1-Pro-Preview (Gemini)\cite{gemini3.1Pro}, all accessed via the OpenRouter API for consistent deployment. 
% Because the GPT and Claude APIs did not support direct video input at the time of evaluation, we evaluated these models only on the image-based SVLAT visualizations.
\hot{For animation items, GPT and Claude were evaluated using frame extraction (1 frame per second), as their APIs did not support direct video input at the time of evaluation.}
The open-source models are Qwen3.5-9B (Qwen)\cite{qwen35blog}, InternVL3.5-8B (InternVL)\cite{chen2024internvl}, and LLaVA-OneVision-1.5-8B-Instruct (LLaVA-OneVision)\cite{li2024llava}.
All models were evaluated with \textit{temperature} set to 0 and \textit{max\_tokens} set to 300 to improve reproducibility and consistency across models and runs. Setting the \textit{temperature} to 0 makes decoding as deterministic as possible, thereby reducing sampling variability in both answer selection and rationale generation. We set \textit{max\_tokens} to 300 to ensure sufficient space for the required JSON output and brief explanation while avoiding unnecessarily long responses. Because some backend nondeterminism may remain, we repeated each evaluation 10 times and report the averaged results.

{\bf Prompt design.}\
We use a single instruction prompt for all models to standardize the evaluation setting. The prompt frames the model as a SciVis expert and explicitly constrains it to use only the provided visualization and caption. It also requires the model to return a structured JSON object containing the selected option, the full answer text, and a brief rationale. To discourage unsupported guessing, the prompt includes a \texttt{"Not sure"} option when the answer cannot be determined confidently from the given evidence. The prompt used in our experiments is shown in Figure \ref{fig:prompt}.
This prompt is designed to align with the closed-world principle of SVLAT and to make model outputs easier to parse automatically.

% \begin{lstlisting}
% PROMPT = """
% You are a scientific visualization expert. You will be given a scientific visualization (image or animation), its caption, and a question about it.

% Provide your response in the following JSON format:
% {
%   "choice": "<the selected answer option (e.g., 'A', 'B', 'C', 'D') or 'Not sure'>",
%   "answer_value": "<the full text of the selected answer option>",
%   "rationale": "<your reason for choosing this answer, in 5 sentences or fewer>"
% }

% Requirements:
% 1. Base your answer strictly on the provided visualization and caption. Do not use prior knowledge or make assumptions.
% 2. Keep your rationale concise, with a maximum of 5 sentences.
% 3. If the answer cannot be determined confidently from the provided visualization and caption, set "choice" to "Not sure" and explain why in your rationale.
% 4. Return valid JSON only. Do not include Markdown or any extra text outside the JSON object.
% """
% \end{lstlisting}

{\bf Evaluation protocol.}\
We evaluate each model on SVLAT one question at a time. For each item, we run the model 10 times using different random seeds and compute the average score across runs. This repeated-run protocol helps reduce the impact of residual sampling variability or backend nondeterminism.
In addition to the selected answer, we record the model's rationale for each response. These explanations are used to examine whether the model's reasoning is grounded in the provided visualization and caption rather than unsupported prior knowledge or speculation. Although the rationale is not used directly for scoring accuracy, it provides qualitative evidence about how the model arrives at its answers and whether incorrect responses stem from misinterpretation of visual evidence, overconfident guessing, or reliance on external knowledge.
We also analyze performance by SciVis techniques and task categories to identify systematic strengths and weaknesses. \hot{To contextualize MLLM capabilities, we compare accuracy against the human performance baseline from the original SVLAT study \cite{Do-arXiv26}, established via an online tryout with 485 non-expert participants (246 females; ages 18–65; education ranging from high school to postgraduate).}

\begin{table}[htb]
\vspace{-0.1in}
\caption{Overall accuracy (mean $\pm$ std) per model split by item format.
Image = static image items; Animation = animation items; All = combined. 
\hot{For GPT-5.4 and Claude-Opus-4.6, animation items were evaluated via frame extraction.}
}
\label{tab:literacy}
\vspace{-0.1in}
\centering
{\fontfamily{pag}\selectfont
\fontsize{6.0}{7.6}\selectfont
\setlength{\tabcolsep}{3pt}
\begin{tabular}{lccc}
\toprule
 & \multicolumn{3}{c}{Accuracy (\%)} \\
\cmidrule(lr){2-4}
Model & \multicolumn{1}{c}{Image} & \multicolumn{1}{c}{Animation} & \multicolumn{1}{c}{All} \\
\midrule
GPT-5.4 & $75.7 \pm 5.5$ & \textcolor{black}{$73.6 \pm 10.4$} & \textcolor{black}{$75.1 \pm 4.9$} \\
Claude-Opus-4.6 & $81.7 \pm 5.8$ & \textcolor{black}{$59.3 \pm 11.7$} & \textcolor{black}{$75.3 \pm 5.5$} \\
Gemini-3.1-Pro-Preview & $90.9 \pm 4.3$ & $82.9 \pm 8.5$ & $88.6 \pm 3.9$ \\
Qwen3.5-9B & $69.4 \pm 6.0$ & $67.1 \pm 9.4$ & $68.8 \pm 5.0$ \\
InternVL3.5-8B & $60.9 \pm 6.7$ & $72.9 \pm 9.9$ & $64.3 \pm 5.6$ \\
LLaVA-OneVision-1.5-8B-Instruct & $60.3 \pm 6.9$ & $73.6 \pm 9.7$ & $64.1 \pm 5.8$ \\
Human & $76.2 \pm 2.4$ & $73.9 \pm 4.7$ & $75.6 \pm 2.2$ \\
\bottomrule
\end{tabular}
}
\vspace{-0.1in}
\end{table}

\vspace{-0.1in}
\section{Results}

\subsection{Overall Performance}

Table \ref{tab:literacy} reports the overall performance of the six evaluated MLLMs across 10 runs and of 485 human participants on SVLAT. Gemini is the strongest model by a wide margin, achieving 90.9\% on image items, 82.9\% on animation items, and 88.6\% overall, exceeding human performance of 76.4\%, 74.5\%, and 75.9\%, respectively.
The three open-weight models trail humans in combined accuracy, with Qwen performing best (68.8\% overall), followed by InternVL (64.3\%) and LLaVA-OneVision (64.1\%). 
% On image-only items, Claude reaches 81.7\%, outperforming both humans and GPT (75.7\%), although neither was evaluated on animation items. On animation items, Gemini remains the strongest model at 82.9\%. 
\hot{On animation items, Gemini remains the top-performing model. Via frame extraction, GPT matches human-level accuracy (73.6\% vs.\ 73.9\%), while Claude scores the lowest of all models and humans at 59.3\%.}
Among the open-source models, LLaVA-OneVision (73.6\%) and InternVL (72.9\%) come close to the human level, and both perform better on animation items than on static image items, whereas Qwen remains lower at 67.1\%. \hot{MLLMs exhibit greater performance heterogeneity across item types (SD: 3.9--11.7) than humans (SD: 2.2--4.7).}

\begin{figure}[!t]
    \centering
    \includegraphics[width=\columnwidth]{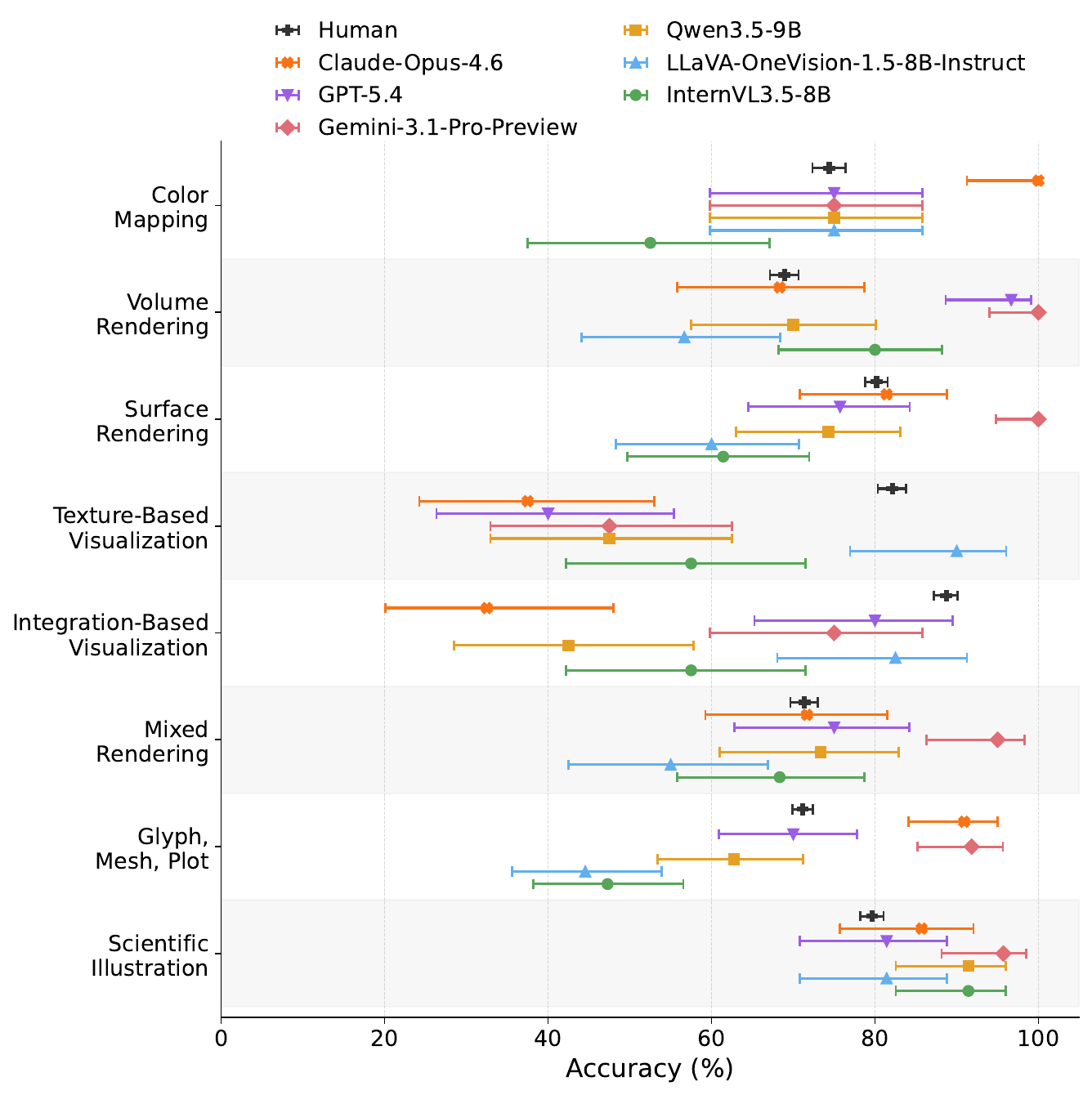}
    \vspace{-0.25in}
    \caption{\hot{Model and human performance across different techniques in SVLAT. Each dot shows the mean accuracy, with the bars indicating 95\% confidence intervals.}
    }
    \label{fig:technique}
   \vspace{-0.2in}
\end{figure}

\vspace{-0.075in}
\subsection{Performance by Techniques}

Human and model performance differ substantially across techniques, as shown in Figure \ref{fig:technique}. Humans exhibit the most balanced pattern, with mean accuracy ranging from 68.9\% to 88.8\% across all techniques, whereas the MLLMs show much larger fluctuations. Among the evaluated models, Gemini is the strongest and broadest-performing system. It achieves 100.0\% on \textsf{Surface Rendering} and \textsf{Volume Rendering}, 95.0\% on \textsf{Mixed Rendering}, 95.7\% on \textsf{Scientific Illustration}, and 91.8\% on \textsf{Glyph, Mesh, Plot}, exceeding human performance on the same techniques. \hot{Closed-source models consistently outperform open-source models across \textsf{Surface Rendering}, \textsf{Mixed Rendering}, and \textsf{Glyph, Mesh, Plot}, with the gap being most pronounced on the latter.}

Across all techniques, \textsf{Scientific Illustration} is the most accessible to humans and all models. Human accuracy reaches 79.5\%, while all models perform strongly, \hot{ranging from 81.4\% to 95.7\%}. In contrast, \textsf{Texture-Based Visualization} and \textsf{Integration-Based Visualization} reveal a clearer gap between human and model performance. Humans achieve high accuracies on both techniques, at 82.0\% and 88.8\%, respectively.
%LLaVA-OneVision is the main exception, performing competitively on \textsf{Texture-Based Visualization} (90.0\%) and \textsf{Integration-Based Visualization} (82.5\%), whereas Gemini, Qwen, and InternVL remain substantially lower on both.
\hot{LLaVA-OneVision is the main exception on \textsf{Texture-Based Visualization} (90.0\%), where all other models fall substantially below it. On \textsf{Integration-Based Visualization}, LLaVA-OneVision (82.5\%), GPT (80.0\%), and Gemini (75\%) remain competitive with humans, whereas other models are notably lower.} 
Per-item accuracies for MLLMs and humans, categorized by techniques, are furnished in Appendix~\ref{sec:app}.

\begin{figure}[!t]
    \centering
    \includegraphics[width=\columnwidth]{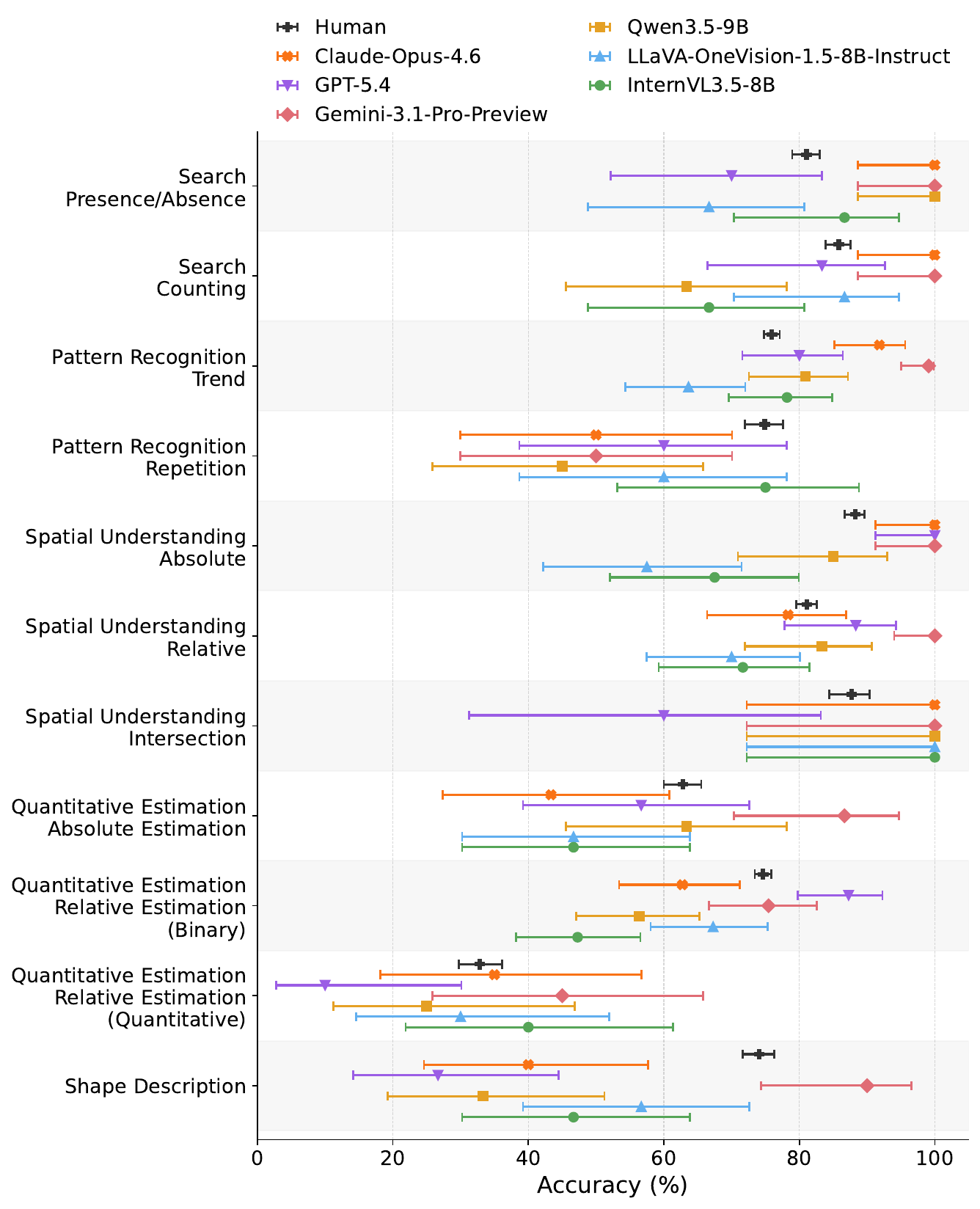}
    \vspace{-0.25in}
    \caption{\hot{Model and human performance across different tasks in SVLAT. Each dot shows the mean accuracy, with the bars indicating 95\% confidence intervals.}
    }
    \label{fig:task}
   \vspace{-0.2in}
\end{figure}

\vspace{-0.075in}
\subsection{Performance by Tasks}

Figure \ref{fig:task} shows that performance varies substantially by task. Across all categories, \textsf{Search and Spatial Understanding} tasks are the most accessible, whereas \textsf{Quantitative Estimation--Relative Estimation (Quantitative)} is the most difficult; indeed, it is the hardest task for every category, with accuracy remaining below 50.0\% throughout. Humans perform particularly well on \textsf{Spatial Understanding--Absolute} (88.1\%), \textsf{Spatial Understanding--Intersection} (87.7\%), and \textsf{Search--Counting} (85.6\%), and the models follow the same broad pattern. 
%Gemini reaches 100.0\% on two \textsf{Search} tasks and three \textsf{Spatial Understanding} tasks, while all three open-source models also reach 100.0\% on \textsf{Spatial Understanding--Intersection}.

Among all models, Gemini performs best, achieving the highest accuracy on nearly every task. The main exception is \textsf{Pattern Recognition--Repetition} \hot{and \textsf{Quantitative Estimation--Relative Estimation (Quantitative)}, where Gemini drops to 50.0\% and 45.0\%, respectively.}
\hot{For \textsf{Spatial Understanding--Absolute}, all three closed-source models max out, whereas open-source models are notably lower.}
%well below humans (74.7\%) and InternVL (75.0\%). 
%In contrast, 
The open-source models are much less competitive with humans. Qwen and InternVL exceed human performance only on \textsf{Search--Presence/Absence} and \textsf{Spatial Understanding--Intersection}, while LLaVA-OneVision exceeds humans only on \textsf{Spatial Understanding--Intersection}. On the remaining tasks, the open-source models perform at or below the human level. \hot{It is worth noting that \textsf{Quantitative Estimation--Relative (Quantitative)}, \textsf{Spatial Understanding--Intersection}, and \textsf{Pattern Recognition--Repetition} are each represented by only one or two items in the current benchmark, so the above findings for these categories should be treated as preliminary.}

% \vspace{-0.05in}
% \subsection{Uncertainty Analysis}

\vspace{-0.075in}
\subsection{Error Analysis}
% The preceding technique- and task-specific results show that MLLM performance on SVLAT is highly uneven, with clear differences not only across techniques and tasks but also between closed-source and open-source models. To better understand the sources of these differences, we qualitatively examine the models' rationales to identify recurring patterns in how they interpret, misinterpret, or fail to ground visual evidence. Because Gemini is the strongest and most broadly capable model in our benchmark, we focus primarily on its behavior, while also contrasting it with the open-source models to clarify which failure modes are shared and which appear more model-specific.
%[I shortened this paragraph to leave more space for the analysis]
\hot{The preceding results show that MLLM performance on SVLAT varies substantially across techniques, tasks, and model families. To better understand the sources of these differences, we qualitatively examine model rationales to identify recurring patterns in how models interpret, misinterpret, or fail to ground visual evidence.}

\begin{figure}[!t]
    \centering
    \includegraphics[width=0.875\columnwidth]{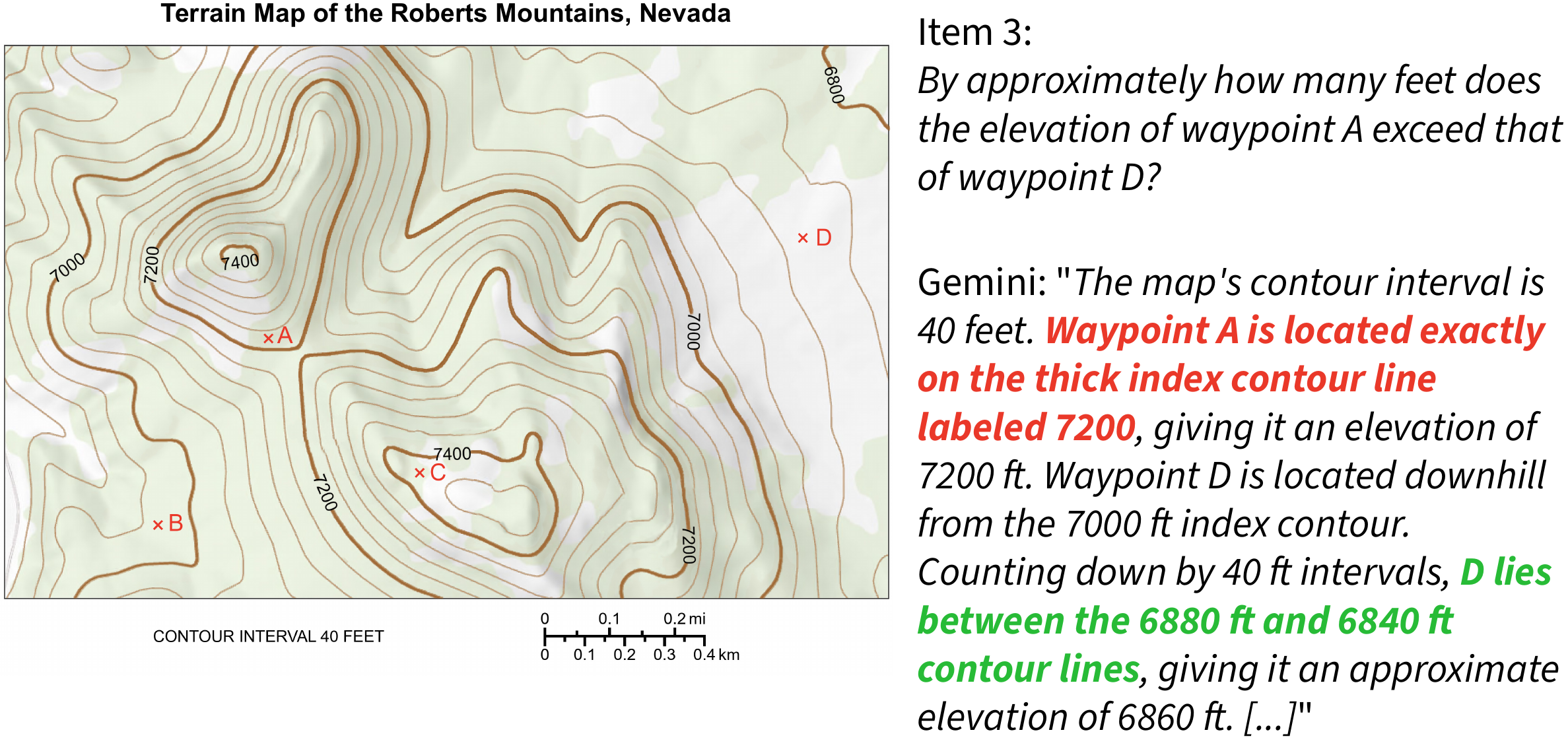}
    \vspace{-0.125in}
    \caption{Example of fine-grained quantitative estimation difficulty. MLLM correctly identified the contour interval and waypoint D, but misread waypoint A as lying on the 7200 ft contour line.}
    \label{fig:item3}
    \vspace{-0.25in}
\end{figure}

\textbf{Fine-grained quantitative misestimation.}\ 
Models struggle with precise quantitative judgment when the answer depends on careful visual estimation rather than coarse pattern recognition. In Item 3, Gemini correctly recognized that ``{\em the map's contour interval is 40 feet}'' and located waypoint D, stating that it ``{\em lies between the 6880 ft and 6840 ft contour lines}'' (see Figure \ref{fig:item3}). However, it misread waypoint A by claiming that it is ``{\em located exactly on the thick index contour line labeled 7200,}'' even though A lies very close to the 7200 contour line while remaining between the 7200 ft and 7240 ft contour lines. A similar problem appears in Item 41 ``{\em What is the approximate radius of the lipid envelope?}'' (refer to Appendix~\ref{sec:app}). \hot{Gemini used the scale bar correctly, but estimated the diameter to be ``{\em approximately five times the length of the scale bar,}'' yielding a radius of 50 nm; Claude produced the same overestimate across all 10 runs, and GPT did so in 3 of 10 runs. By contrast, the open-source models showed no consistent bias, producing radius estimates ranging from 30 to 50 nm across runs. The correct diameter is approximately 4 times the 20 nm scale bar, giving a radius of about 40 nm. These examples suggest that the models can often identify the relevant quantitative encoding and coarse measurement strategy, a weakness reflected in the particularly low MLLM scores on \textsf{Quantitative Estimation--Relative Estimation (Quantitative)}.}

\begin{figure}[htb]
%    \vspace{-0.1in}
    \centering
    \includegraphics[width=0.9\columnwidth]{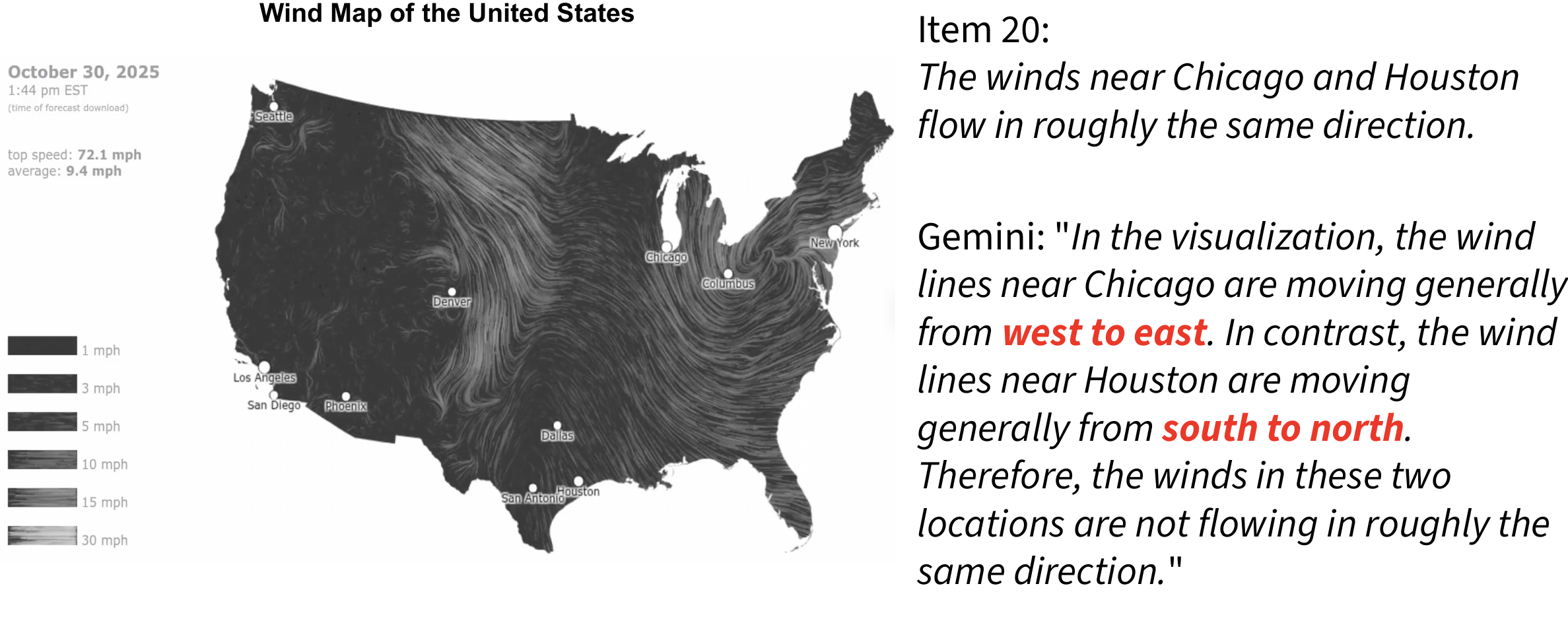}
    \vspace{-0.125in}
    \caption{Example of flow-direction interpretation error in a texture-based visualization. MLLM misread the local wind directions near Chicago and Houston, which are both roughly northwest to southeast.}
    \label{fig:item20}
   \vspace{-0.1in}
\end{figure}

\textbf{Flow-direction interpretation error.}\ 
Texture-based visualizations exposed difficulty in decoding local flow direction from animated vector fields. 
\hot{In Item 20, Gemini and Qwen incorrectly concluded that the winds near Chicago and Houston flow in different directions, reasoning that Chicago moves ``{\em from west to east}'' while Houston moves ``{\em from south to north.}''}
In fact, the winds in both regions flow in roughly the same direction, approximately from northwest to southeast (see Figure \ref{fig:item20}). 
\hot{This pattern of difficulty extends even to models that answered correctly: InternVL identified the same direction at both cities but misread it as ``{\em west to east,}'' while LLaVA avoided any specific direction, stating only that arrows ``{\em point in a similar direction.}''}
This suggests that the model misread local motion in the animated flow field, possibly by confusing streak orientation with the actual flow direction. \hot{Because most models correctly identified the hurricane's direction in Item 9 (refer to Appendix~\ref{sec:app}), the difficulty appears specific to dense texture-based motion decoding rather than to directional reasoning more generally.}

% \begin{figure}[!t]
%     \centering
%     \includegraphics[width=\columnwidth]{figures/item20.pdf}
%     \vspace{-0.15in}
%     \caption{Example of flow-direction interpretation error in a texture-based visualization. MLLM misread the local wind directions near Chicago and Houston, which in fact are both roughly northwest to southeast.}
%     \label{fig:item20}
% \end{figure}

\begin{figure}[!t]
    \centering
    \includegraphics[width=0.9\columnwidth]{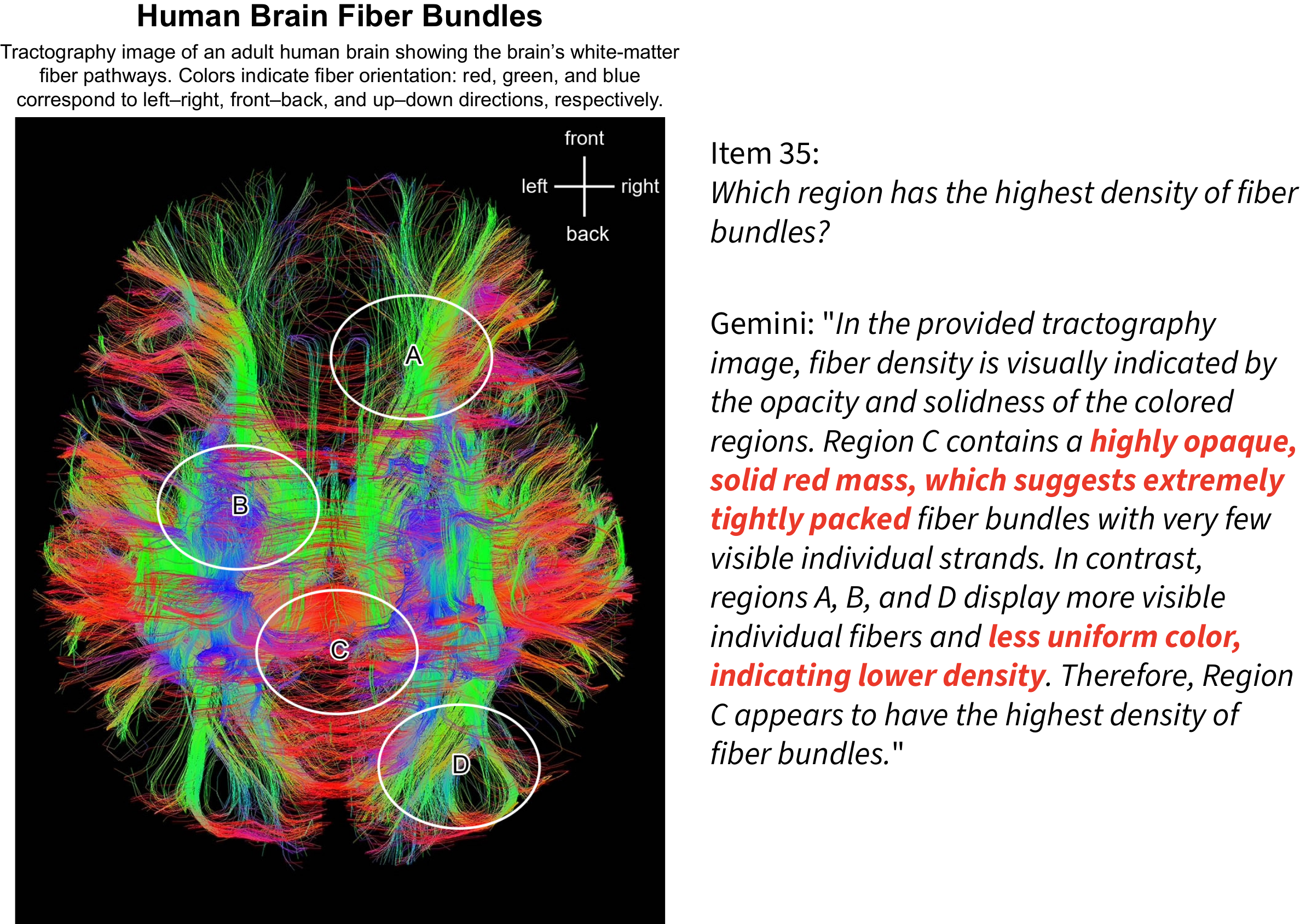}
    \vspace{-0.075in}
    \caption{Example of an encoding-mapping error during higher-level SciVis interpretation.}
    \label{fig:item35}
    \vspace{-0.25in}
\end{figure}

\textbf{Encoding-mapping error during higher-level SciVis interpretation.}\ 
Some failures occur when the model makes a higher-level interpretation, but grounds it in an unsupported visual mapping. In Item 35, Gemini inferred that fiber density was indicated by the opacity and solidness of the colored regions, and on that basis concluded that region C had the highest density because it appeared as a ``{\em highly opaque, solid red mass}'' (see Figure \ref{fig:item35}). \hot{Claude reached the same wrong conclusion across all 10 runs, but via a different unsupported mapping: it invoked anatomical prior knowledge, stating that region C ``{\em corresponds to the splenium of the corpus callosum ... densely packed with fibers of multiple orientations (red, green, and blue),}'' thereby grounding its density judgment in imported neuroanatomical knowledge and color variety rather than actual fiber concentration in the image.} The item requires a SciVis judgment on fiber density, which can be estimated from the overall concentration and fiber overlap. Instead, \hot{both models} justified their answer with unsupported cues
%, such as apparent solidity and the region C's salient red appearance, 
even though the caption specifies that color encodes fiber orientation. The error, therefore, is not simply that the model made a higher-level inference, but that it grounded this inference in an unsupported encoding rather than in the actual evidence provided by the visualization and caption.

\vspace{-0.1in}
\section{Conclusions and Future Work}
Our results show that current MLLMs do not yet exhibit uniform SciVis literacy. Performance varies substantially across techniques and tasks. Across techniques, the models perform best on \textsf{Scientific Illustration}, while Gemini also performs strongly on \textsf{Surface Rendering}, \textsf{Volume Rendering}, and \textsf{Mixed Rendering}. Across tasks, their strongest results are in \textsf{Search} and \textsf{Spatial Understanding}, particularly in \textsf{Presence/Absence}, \textsf{Counting}, and \textsf{Intersection}. However, the open-source models remain less robust overall and show narrower, more uneven strengths. The qualitative analysis shows that they reflect recurring weaknesses in fine-grained quantitative estimation, flow-direction interpretation in texture-based visualizations, and unsupported encoding inference during higher-level SciVis interpretation. These patterns suggest that current MLLMs can often identify the general structure of a task, but still struggle when success depends on precise measurement, decoding local motion in dense dynamic fields, or grounding higher-level judgments in the actual encoding specified by the visualization and caption.

This study has several limitations. \hot{GPT and Claude were evaluated on animation items via frame extraction, as their APIs did not support direct video input during the evaluation.} \hot{In addition, the evaluation is based on a single benchmark, SVLAT, and a single standardized prompt format, and the open-source comparison is limited to lightweight models.} \hot{Additionally, some tasks are represented by only one or two items, which limits the reliability of per-task estimates and may not fully capture the range of difficulty within those tasks.} Future work should extend this benchmark to larger and more diverse models, richer prompting and tool-use settings, and additional SciVis-native evaluation instruments. Overall, our findings show that current MLLMs are selectively capable rather than broadly SciVis literate. SciVis literacy is, therefore, a necessary benchmark dimension because scientific visualizations require forms of spatial, temporal, and encoding-grounded interpretation that are not captured by general multimodal evaluations.

\vspace{-0.1in}
\section*{Figure Credits}
Figure~\ref{fig:item3}: Terrain Map of the Roberts Mountains, Nevada---Image adapted from USGS National Geologic Map Database.
Figure~\ref{fig:item20}: Wind Map of the United States---Animation from hint.fm.
Figure~\ref{fig:item35}: Human Brain Fiber Bundles---Image adapted from Zeynep Saygin / MIT Koch Institute.

\vspace{-0.075in}
\acknowledgments{This research was supported in part by the U.S.\ National Science Foundation through grants IIS-2101696, OAC-2104158, IIS-2401144, and CCF-2550610. The authors thank the anonymous reviewers for their insightful comments.}

\bibliographystyle{abbrv-doi-hyperref-narrow}

\vspace{-0.05in}
%\bibliography{refs}
\bibliography{refs-abbv}

\appendix % You can use the `hideappendix` class option to skip everything after \appendix
\crefalias{section}{appendix} % this is to make sure that cleverref switches to referring to Appx. X from here on

\clearpage

\setcounter{section}{0}
\setcounter{figure}{0}
\setcounter{table}{0}
\setcounter{page}{1}

\onecolumn

\section{Additional Details and Results}
\label{sec:app}

% Switch back to two columns for the rest of the paper
%\clearpage
%\twocolumn

\begin{figure}[htb]
%\vspace{-0.1in}
\centering
\includegraphics[width=0.9\linewidth]{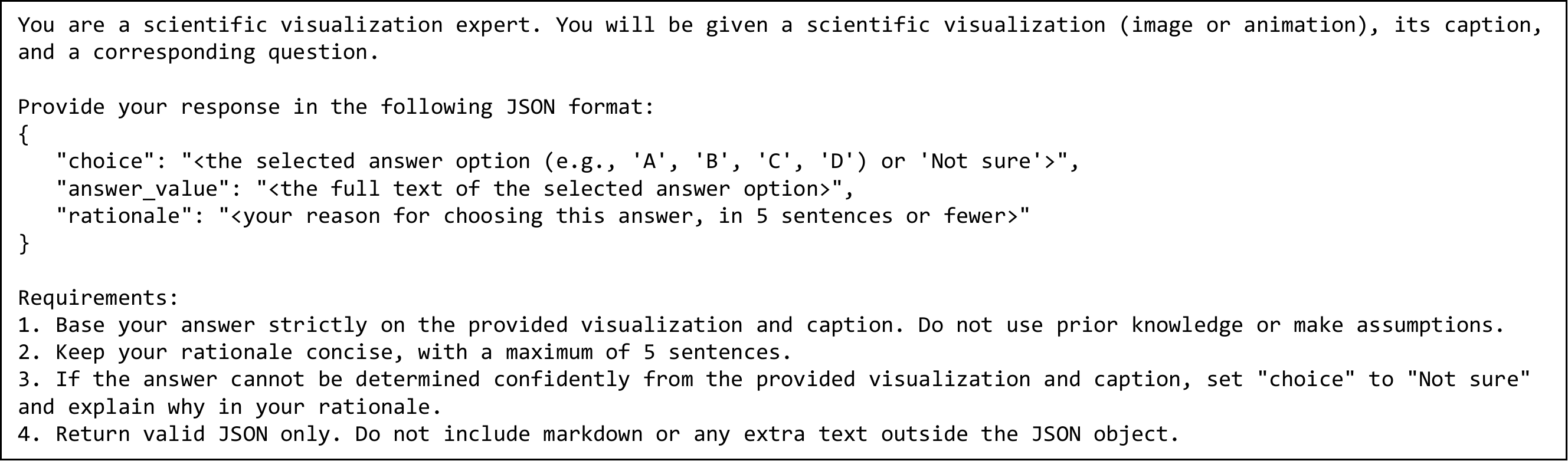}
\vspace{-0.1in}
\caption{The prompt template for our MLLM literacy experiments.}
\label{fig:prompt}
%\vspace{-0.1in}
\end{figure}

\begin{table}[!h]
\centering
\setlength{\tabcolsep}{4pt}
\caption{Per-item accuracies for all MLLMs and humans, categorized by techniques. I: static image, A: animation. \hot{GPT and Claude were evaluated on animation items via frame extraction.}}
\label{tab:per-item-results}
\vspace{-0.1in}

{\fontfamily{pag}\selectfont\fontsize{5.0}{6.4}\selectfont
\begin{tabularx}{\textwidth}{@{}>{\raggedright\arraybackslash}m{1.8cm}%
>{\raggedright\arraybackslash}p{0.65cm}%
>{\centering\arraybackslash}p{0.5cm}%
>{\raggedright\arraybackslash}X%
>{\centering\arraybackslash}p{0.65cm}%
>{\centering\arraybackslash}p{1.2cm}%
>{\centering\arraybackslash}p{0.8cm}%
>{\centering\arraybackslash}p{1.05cm}%
>{\centering\arraybackslash}p{1.5cm}%
>{\centering\arraybackslash}p{1.25cm}%
>{\centering\arraybackslash}p{0.6cm}@{}}
\toprule
\textbf{Technique} & \textbf{Item ID} & \textbf{Format} & \textbf{Task} & \textbf{GPT-5.4} & \textbf{Gemini-3.1-Pro-Preview} & \textbf{Claude-Opus-4.6} & \textbf{Qwen3.5-9B} & \textbf{LLaVA-OneVision-1.5-8B} & \textbf{InternVL3.5-8B} & \textbf{Human} \\
\midrule
\multirow[c]{4}{2.7cm}{\makecell[l]{Color Mapping}} & Item 13 & I & Quantitative Estimation - Absolute Estimation & 100.0 & 90.0 & 100.0 & 100.0 & 0.0 & 0.0 & 72.2 \\
 & Item 14 & I & Quantitative Estimation - Relative Estimation (Binary) & 30.0 & 10.0 & 100.0 & 0.0 & 100.0 & 20.0 & 56.5 \\
 & Item 15 & I & Search - Presence/Absence & 100.0 & 100.0 & 100.0 & 100.0 & 100.0 & 100.0 & 77.9 \\
 & Item 16 & I & Pattern Recognition - Trend & 70.0 & 100.0 & 100.0 & 100.0 & 100.0 & 90.0 & 91.2 \\
\midrule
\multirow[c]{6}{2.7cm}{\makecell[l]{Volume Rendering}} & Item 47 & I & Quantitative Estimation - Relative Estimation (Binary) & 100.0 & 100.0 & 90.0 & 70.0 & 100.0 & 100.0 & 70.0 \\
 & Item 54 & A & Pattern Recognition - Trend & \textcolor{black}{80.0} & 100.0 & \textcolor{black}{10.0} & 50.0 & 0.0 & 80.0 & 51.7 \\
 & Item 55 & A & Quantitative Estimation - Relative Estimation (Binary) & \textcolor{black}{100.0} & 100.0 & \textcolor{black}{100.0} & 100.0 & 100.0 & 100.0 & 65.6 \\
 & Item 56 & A & Pattern Recognition - Trend & \textcolor{black}{100.0} & 100.0 & \textcolor{black}{100.0} & 100.0 & 90.0 & 100.0 & 71.7 \\
 & Item 66 & A & Quantitative Estimation - Relative Estimation (Binary) & \textcolor{black}{100.0} & 100.0 & \textcolor{black}{10.0} & 0.0 & 0.0 & 0.0 & 68.4 \\
 & Item 67 & A & Pattern Recognition - Trend & \textcolor{black}{100.0} & 100.0 & \textcolor{black}{100.0} & 100.0 & 50.0 & 100.0 & 84.7 \\
\midrule
\multirow[c]{7}{2.7cm}{\makecell[l]{Surface Rendering}} & Item 25 & I & Search - Counting & 50.0 & 100.0 & 100.0 & 20.0 & 80.0 & 0.0 & 89.1 \\
 & Item 27 & I & Shape Description & 30.0 & 100.0 & 0.0 & 0.0 & 0.0 & 30.0 & 72.0 \\
 & Item 28 & I & Spatial Understanding - Intersection & 60.0 & 100.0 & 100.0 & 100.0 & 100.0 & 100.0 & 87.7 \\
 & Item 31 & I & Spatial Understanding - Relative & 100.0 & 100.0 & 70.0 & 100.0 & 0.0 & 0.0 & 73.4 \\
 & Item 32 & I & Spatial Understanding - Relative & 100.0 & 100.0 & 100.0 & 100.0 & 40.0 & 100.0 & 88.4 \\
 & Item 51 & A & Pattern Recognition - Trend & \textcolor{black}{90.0} & 100.0 & \textcolor{black}{100.0} & 100.0 & 100.0 & 100.0 & 86.8 \\
 & Item 52 & A & Quantitative Estimation - Relative Estimation (Binary) & \textcolor{black}{100.0} & 100.0 & \textcolor{black}{100.0} & 100.0 & 100.0 & 100.0 & 63.2 \\
\midrule
\multirow[c]{4}{2.7cm}{\makecell[l]{Texture-Based}\\{Visualization}} & Item 17 & A & Pattern Recognition - Trend & \textcolor{black}{100.0} & 100.0 & \textcolor{black}{100.0} & 100.0 & 90.0 & 100.0 & 82.7 \\
 & Item 18 & A & Quantitative Estimation - Relative Estimation (Binary) & \textcolor{black}{40.0} & 20.0 & \textcolor{black}{0.0} & 30.0 & 100.0 & 20.0 & 79.0 \\
 & Item 19 & A & Shape Description & \textcolor{black}{0.0} & 70.0 & \textcolor{black}{50.0} & 20.0 & 70.0 & 10.0 & 83.6 \\
 & Item 20 & A & Pattern Recognition - Repetition & \textcolor{black}{20.0} & 0.0 & \textcolor{black}{0.0} & 40.0 & 100.0 & 100.0 & 83.2 \\
\midrule
\multirow[c]{4}{2.7cm}{\makecell[l]{Integration-Based}\\{Visualization}} & Item 33 & I & Pattern Recognition - Trend & 100.0 & 100.0 & 100.0 & 60.0 & 100.0 & 100.0 & 85.8 \\
 & Item 35 & I & Quantitative Estimation - Relative Estimation (Binary) & 90.0 & 0.0 & 0.0 & 10.0 & 40.0 & 0.0 & 88.3 \\
 & Item 57 & I & Quantitative Estimation - Relative Estimation (Binary) & 100.0 & 100.0 & 30.0 & 100.0 & 100.0 & 100.0 & 96.9 \\
 & Item 59 & I & Spatial Understanding - Relative & 30.0 & 100.0 & 0.0 & 0.0 & 90.0 & 30.0 & 83.0 \\
\midrule
\multirow[c]{6}{2.7cm}{\makecell[l]{Mixed Rendering}} & Item 60 & I & Spatial Understanding - Absolute & 100.0 & 100.0 & 100.0 & 100.0 & 0.0 & 100.0 & 87.1 \\
 & Item 61 & I & Shape Description & 50.0 & 100.0 & 70.0 & 80.0 & 100.0 & 100.0 & 66.1 \\
 & Item 62 & I & Quantitative Estimation - Relative Estimation (Binary) & 100.0 & 100.0 & 100.0 & 60.0 & 0.0 & 0.0 & 54.1 \\
 & Item 68 & A & Pattern Recognition - Trend & \textcolor{black}{100.0} & 100.0 & \textcolor{black}{100.0} & 100.0 & 100.0 & 100.0 & 95.3 \\
 & Item 70 & A & Quantitative Estimation - Relative Estimation (Binary) & \textcolor{black}{100.0} & 100.0 & \textcolor{black}{60.0} & 50.0 & 100.0 & 70.0 & 92.2 \\
 & Item 71 & A & Quantitative Estimation - Relative Estimation (Quantitative) & \textcolor{black}{0.0} & 70.0 & \textcolor{black}{0.0} & 50.0 & 30.0 & 40.0 & 27.0 \\
\midrule
\multirow[c]{11}{2.7cm}{\makecell[l]{Glyph, Mesh, Plot}} & Item 1 & I & Quantitative Estimation - Absolute Estimation & 10.0 & 100.0 & 30.0 & 50.0 & 100.0 & 100.0 & 71.2 \\
 & Item 2 & I & Quantitative Estimation - Relative Estimation (Binary) & 100.0 & 100.0 & 100.0 & 100.0 & 0.0 & 10.0 & 83.7 \\
 & Item 3 & I & Quantitative Estimation - Relative Estimation (Quantitative) & 20.0 & 20.0 & 70.0 & 0.0 & 30.0 & 40.0 & 38.8 \\
 & Item 4 & I & Pattern Recognition - Trend & 70.0 & 90.0 & 100.0 & 60.0 & 0.0 & 20.0 & 55.7 \\
 & Item 5 & I & Pattern Recognition - Trend & 60.0 & 100.0 & 100.0 & 70.0 & 70.0 & 40.0 & 61.5 \\
 & Item 7 & I & Search - Presence/Absence & 100.0 & 100.0 & 100.0 & 100.0 & 60.0 & 60.0 & 74.8 \\
 & Item 9 & I & Spatial Understanding - Absolute & 100.0 & 100.0 & 100.0 & 90.0 & 100.0 & 40.0 & 87.6 \\
 & Item 10 & I & Spatial Understanding - Absolute & 100.0 & 100.0 & 100.0 & 50.0 & 30.0 & 30.0 & 87.5 \\
 & Item 12 & I & Pattern Recognition - Trend & 10.0 & 100.0 & 100.0 & 50.0 & 0.0 & 30.0 & 66.8 \\
 & Item 22 & I & Search - Counting & 100.0 & 100.0 & 100.0 & 70.0 & 80.0 & 100.0 & 84.2 \\
 & Item 23 & I & Pattern Recognition - Repetition & 100.0 & 100.0 & 100.0 & 50.0 & 20.0 & 50.0 & 65.5 \\
\midrule
\multirow[c]{7}{2.7cm}{\makecell[l]{Scientific}\\{Illustration}} & Item 36 & I & Spatial Understanding - Relative & 100.0 & 100.0 & 100.0 & 100.0 & 100.0 & 100.0 & 90.0 \\
 & Item 37 & I & Spatial Understanding - Relative & 100.0 & 100.0 & 100.0 & 100.0 & 100.0 & 100.0 & 72.4 \\
 & Item 40 & I & Spatial Understanding - Relative & 100.0 & 100.0 & 100.0 & 100.0 & 90.0 & 100.0 & 79.5 \\
 & Item 41 & I & Quantitative Estimation - Absolute Estimation & 60.0 & 70.0 & 0.0 & 40.0 & 40.0 & 40.0 & 43.1 \\
 & Item 43 & I & Search - Counting & 100.0 & 100.0 & 100.0 & 100.0 & 100.0 & 100.0 & 84.3 \\
 & Item 45 & I & Search - Presence/Absence & 10.0 & 100.0 & 100.0 & 100.0 & 40.0 & 100.0 & 90.2 \\
 & Item 46 & I & Spatial Understanding - Absolute & 100.0 & 100.0 & 100.0 & 100.0 & 100.0 & 100.0 & 90.7 \\
\bottomrule
\end{tabularx}
}
\end{table}

\end{document}